# DIAGNOSIS OF DIABETIC RETINOPATHY USING MACHINE LEARNING & DEEP LEARNING TECHNIQUE

**Mr. Eric Shah, Mr. Jay Patel, Mr.Vishal Katheriya, Mr. Parth Pataliya** Department Of Computer Science Gujarat University, Ahmedabad

Abstract
Fundus images are widely used for diagnosing various eye diseases, such as diabetic retinopathy, glaucoma, and age-related macular degeneration. However, manual analysis of fundus images is time-consuming and prone to errors. In this report, we propose a novel method for fundus detection using object detection and machine learning classification techniques. We use a YOLO_V8 to perform object detection on fundus images and locate the regions of interest (ROIs) such as optic disc, optic cup and lesions. We then use machine learning SVM classification algorithms to classify the ROIs into different DR stages based on the presence or absence of pathological signs such as exudates, microaneurysms, and haemorrhages etc. Our method achieves 84% accuracy and efficiency for fundus detection and can be applied for retinal fundus disease triage, especially in remote areas around the world.

Introduction
What is Diabetic Retinopathy?
Diabetic retinopathy (DR) is a common complication of diabetes that affects the retina, which is the light- sensitive tissue at the back of the eye. DR creates lesions in the retina that can damage the vision and lead to total blindness if left untreated. DR is one of the leading causes of blindness among working-age adults worldwide.

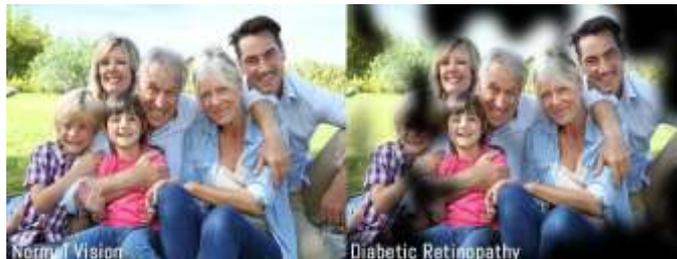
Figure 1:Normal & DR Vision Eye Example

Why is it important to detect DR?
It is important to detect diabetic retinopathy early because early detection and treatment can prevent vision loss and even reverse the damage caused by the condition. Regular eye exams are crucial for catching diabetic retinopathy early, before it progresses to a more advanced stage. Even if you think your diabetes is well controlled, it's still important to attend your annual diabetic eye screening appointment, as this can detect signs of a problem before you notice anything is wrong. Early detection of retinopathy increases the chances of treatment being effective and stopping it from getting worse .

Type Of lesion Diabetic Retinopathy

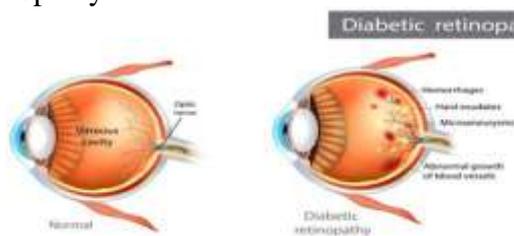
Figure 2:Type Lesion in DR Eye

1. Microaneurysms (Ma's)

Ma's are the first lesions appearing in diabetic retinopathy. Ma's are physical dilations of the smallest intra-retinal blood vessels called capillaries. These lesions appear as small circular, red dots having distinct margins and are no larger than a blood vessel width at the disk margin.

2. Retinal Hemorrhages (H's)

These lesions represent actual bleeding within the retina, and either are a result of ruptured Ma's or when the above-mentioned capillaries become leaky enough to let blood out of the blood vessels (just as when we get a cut on our skin). These hemorrhages can be a variety of shapes including dot, blot and flame-shaped. They are usually larger than Ma's (equal to or larger than a blood vessel at the disk margin), with uneven/indistinct edges and coloring.

3. Hard Exudates (HE's)

These are white/yellow cholesterol deposits that usually originate from leaking Ma's. HE's are irregularly shaped, vary in size, are hard edged and often have a "fatty" appearance. HE can, and often are associated with fluid accumulation (retinal edema) within the retina.

4. Soft Exudates (SE)

aka "Cotton Wool Spots" - these lesions appear as white, feathery, fluffy or "cottony" spots. SE's physically represent infarcts, or closures of capillaries, within the retina; however the physical locations of these lesions are in the very exterior layer of the retina (in the nerve fiber layer). Cotton Wool Spots often occur in association with IRMA (discussed below).

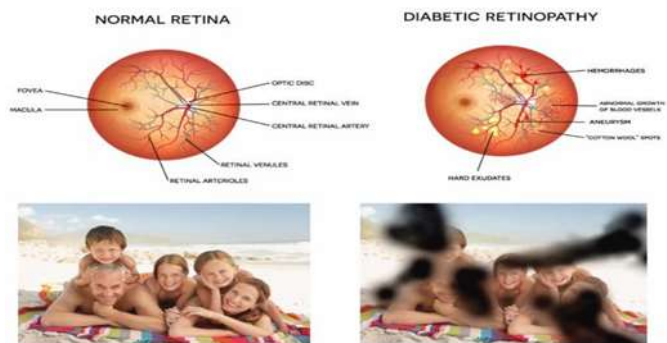

Figure 3: Normal VS Diabetic Retinopathy Eye

5. Intraretinal Microvascular Anomalies (IRMA)

Occurring in the mid to late stages of non- proliferative retinopathy, these appear as spidery abnormal vessels that appear within the retina. They are typically contorted in appearance with sharp corners, often crossing over themselves and they normally do not cross over major veins or arteries. It is thought that IRMA physically represents either dilation of pre-existing capillaries or actual growth of new blood vessels within the retina.

6. Venous Beading (VB)

Also occurring in the late stages of non-proliferative disease, VB occurs when the walls of major retinal veins loose their normal parallel alignment and begin to appear more like a string of sausages. Physically, IRMA represents a situation in which vein walls loose their elasticity and localized areas begin to dilate. This lesion is one of the strongest predictors of progression to proliferative disease. The above standard photograph represents this phenomenon very well.

7. Proliferative Lesions

Proliferative lesions are advanced signs of diabetic retinopathy and in general, require treatment as well as close follow up. The aim of treatment in this stage of disease to reduce the retina's need for oxygen and nutrients, which will cause these proliferative lesions to regress and prevent severe vision loss.

Existing System:
• Diabetic retinopathy is best diagnosed with a comprehensive dilated eye exam. For this exam, drops are placed in your eyes to widen (dilate) your pupils to allow your doctor a better view inside your eyes. The drops can cause your close vision to blur until they wear off, several hours later .

- During the exam, your eye doctor will look for abnormalities in the inside and outside parts of your eyes. There are also other tests that can be used to diagnose diabetic retinopathy, such as fluorescein angiography and optical coherence tomography (OCT) .

Proposed System:
- We used yolo, a state-of-the-art object detection system, to detect DR lesions in retinal images.
- We then used a Support Vector Machine (SVM) to classify the lesions into different stages of DR severity.

Literature Survey

Diabetic retinopathy, nephropathy and neuropathy in patients with type 1 diabetes (T1D) are microvascular complications that can adversely impact disease prognosis and incur greater healthcare costs. Early identification of patients at risk of these microvascular complications using predictive models through machine learning (ML) can be helpful in T1D management. The objective of current review was to systematically identify and summarize published predictive models that used ML to assess the risk of diabetic nephropathy, retinopathy and neuropathy in T1D patients. r results, the problem was that it was quite slow. [1]

Kulwinder Mann and Sukhpreet Kaur worked a lot on segmenting of the blood veins in the retina using the concepts of ANN to detect the disease at an early stage in their research paper in 28. This paper was considered as the base paper from which the motivation was obtained to do research in this exciting field of biomedical engineering. The authors used the supervisory methods for detection purposes and finally they arrived at a conclusion that using ANN they could get very good results & thetraining of the NNsalong with the algorithm was used to detect the DR at an early stage in the human beings, but they worked on only for a limited amount of images taken from the standard database. [2]

The authorsGeethaRamaniet.al.1devised some novel methods for the analysis of retina images. This used both the IP & DM concepts. The team lead by Tang et.al. 2developed a novel method of supervisory classification that was dependent on the filtering of the blood vessels using the Gabor Wavelet Transforms. Next, the team of Abdallah et.al. 3proposed novel methods of segmenting of the blood vessels which were of the same size, i.e., similar dimension. The research team led by Gang et.al. 141523 enunciated the Gaussian Curve was similar to the overview of the retinal blood capillaries and they used it for detection of the retina blood vessels. The authors proposed Gaussian Filtering scheme which was based on modified 2nd order Filter 1213. [3]

Larsen et.al.worked on the detecting of the red colours (lessions) in the diabetic fundus image for detecting the DR disease in his research paper in 31, which was an automated method. Mendoncet.al. did extensive research on segmenting of blood capillaries in the image which used the combination of the centre line detection & reconstruction based on morphological features in 30.

Garciaet.al.proposed a neural network-based detection of hard exudates in retinal images in their research paper in 29. An automated lung module detection which used the matching of the profiles & the BP algo in ANN was devised by Lo&Freedmanin
27. [4]

A concise audit of the imaging techniques for recognizing of the DR disease & its severity effects along with its classification techniques was presented by Madhura &Kakatkar in 33. This paper gave a review of the various strategies for DR identification and grouping into various stages dependent on the extremity levels & furthermore, different image DB's utilized for conducting the research were also presented. Adarsh and Jeyakumari worked on the classification issues, i.e., on the multi-class support vector machine's automatic detection & analysis of the DR disease in
34. Their classification even though it gave very good results, but couldn't satisfy if other types of classifiers were used for the same set of images that were taken from the standard databases. Javeria Aminet.al. carried out a extensive examination of the developments in the recent years w.r.t. the detection of DR disease in their survey paper in 40, which provided as a ready reckoner for the DB researchers. [5]

A hybrid methodology which was depended on the extraction of the features for blood vessel detection was developed by the team of Aslaniet.al.in their research paper in 5. Preparation of text in the pattern of dictionary for classifying the blood veins/capillaries was given by Zhang et.alin their research paper presented in 6. Mathematical models along with clustering using k-means for segmenting of the blood capillaries was proposed by Hassanaet.al. in7. An identification scheme which was based on the characterization of the blood veins (including the shapes, areas, unwanted regional volumes near the retina) was put forward in their research paper by Sinthanayothinet.al.in8. In fact, ant bee-colony optimisation & FC-means cluster method was used by Vermeer et.al. in9 for detecting the fine & coarse blood capillaries, from which they could detect the disease. [6]

Method And Materials
What is Machine Learning?
In the real world, we are surrounded by humans who can learn everything from their experiences with their learning capability, and we have computers or machines which work on our instructions. But can a machine also learn from experiences or past data like a human does? So here comes the role of Machine Learning.
Machine learning enables a machine to automatically learn from data, improve performance from experiences, and predict things without being explicitly programmed. With the help of sample historical data, which is known as training data, machine learning algorithms build a mathematical model that helps in making predictions or decisions without being explicitly programmed. Machine learning brings computer science and statistics together for creating predictive models. Machine learning constructs or uses the algorithms that learn from historical data. The more we will provide the information, the higher will be the performance.
-　　　-A machine has the ability to learn if it can improve its performance by gaining more data.

How does Machine Learning work.

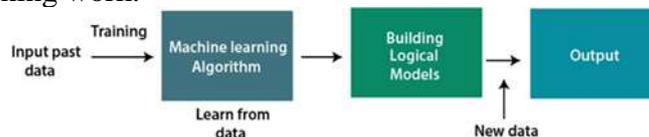

Figure 4 Machine Learning Work Flow Example

Suppose we have a complex problem, where we need to perform some predictions, so instead of writing a code for it, we just need to feed the data to generic algorithms, and with the help of these algorithms, machine builds the logic as per the data and predict the output. Machine learning has changed our way of
thinking about the problem. The below block

diagram explains the working of Machine Learning algorithm:
Type of Machine Learning
•	Supervised Learning:

Supervised learning is a type of machine learning method in which we provide sample labelled data to the machine learning system in order to train it, and on that basis, it predicts the output.
1.	Classification
2.	Regression

•	Unsupervised Learning:
Unsupervised learning is a learning method in which a machine learns without any supervision
1.	Clustering
2.	Association

Support Vector Machine (Supervised Learning classification algorithm)
- Support Vector Machine or SVM is one of the most popular Supervised Learning algorithms, which is used for Classification as well as Regression problems. However, primarily, it is used for Classification problems in Machine Learning.
- The goal of the SVM algorithm is to create the best line or decision boundary that can segregate n-dimensional space into classes so that we can easily put the new data point in the correct category in the future. This best decision boundary is called a hyperplane.
- SVM chooses the extreme points/vectors that help in creating the hyperplane. These extreme cases are called as support vectors, and hence algorithm is termed as Support Vector Machine. Consider the below diagram in which there are two different categories that are classified using a decision boundary or hyperplane:

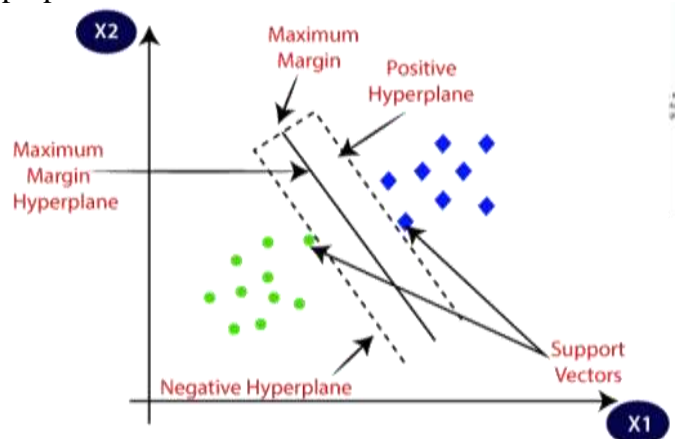

Figure 5:SVM Classification Diagram

Type of SVM
1. Linear SVM: Linear SVM is used for linearly separable data, which means if a dataset can be classified into two classes by using a single straight line, then such data is termed as linearly separable data, and classifier is used called as Linear SVM classifier.
2. Non-linear SVM: Non-Linear SVM is used for non-linearly separated data, which means if a dataset cannot be classified by using a straight line, then such data is termed as non-linear data and classifier used is called as Non-linear SVM classifier.

What is Deep Learning?
Deep learning is a subset of machine learning, which is essentially a neural network with three or more layers. These neural networks attempt to simulate the behaviour of the human brain albeit far from matching its ability allowing it to "learn" from large amounts of data. While a neural network with a single layer can still make approximate predictions, additional hidden layers can help to optimize and refine for accuracy.
How deep learning works?
Deep learning neural networks, or artificial neural networks, attempts to mimic the human brain through a combination of data inputs, weights, and bias. These elements work together to accurately recognize, classify, and describe objects within the data.
Deep neural networks consist of multiple layers of interconnected nodes, each building upon the previous layer to refine and optimize the prediction or categorization. This progression of computations through the network is called forward propagation. The input and output layers of a deep neural network are called visible layers. The input layer is where the deep learning model ingests the data for processing, and the output layer is where the final prediction or classification is made.

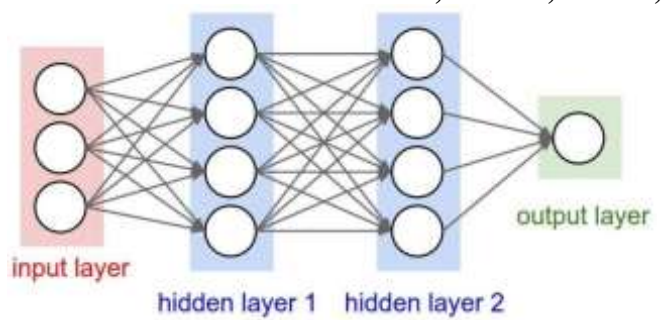
Figure 6 Deep Learning

Another process called backpropagation uses algorithms, like gradient descent, to calculate errors in predictions and then adjusts the weights and biases of the function by moving backwards through the layers in an effort to train the model. Together, forward propagation and backpropagation allow a neural network to make predictions and correct for any errors accordingly. Over time, the algorithm becomes gradually more accurate.

The above Figure 7 describes the simplest type of deep neural network in the simplest terms. However, deep learning algorithms are incredibly complex, and there are different types of neural networks to address specific problems or datasets.

Convolution neural network (CNNs)
A Convolutional Neural Network (CNN) is a type of deep learning algorithm that is particularly well- suited for image recognition and processing tasks. It is made up of multiple layers, including convolutional layers, pooling layers, and fully connected layers.
The convolutional layers are the key component of a CNN, where filters are applied to the input image to extract features such as edges, textures, and shapes. The output of the convolutional layers is then passed through pooling layers, which are used to down- sample the feature maps, reducing the spatial dimensions while retaining the most important information. The output of the pooling layers is then passed through one or more fully connected layers, which are used to make a prediction or classify the image

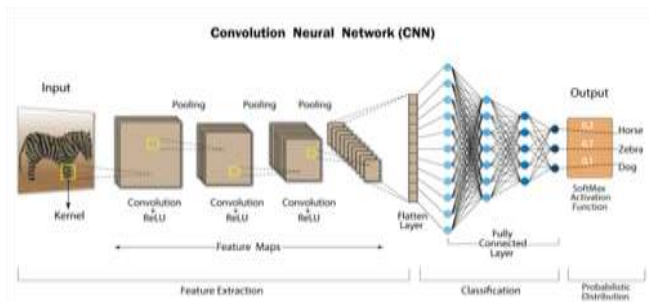
Figure 7 CNN

What is YOLO?
The YOLO (You Only Look Once) algorithm is a popular object detection method in the field of machine learning. It is known for its real-time performance and accuracy in detecting objects in images and videos. Here's a general overview of the methodology behind YOLO:
What is YOLOv8?
YOLOv8 is the newest state-of-the-art YOLO model that can be used for object detection, image classification, and instance segmentation tasks.
YOLOv8 was developed by Ultralytics, who also created the influential and industry-defining YOLOv5 model. YOLOv8 includes numerous architectural and developer experience changes and improvements over YOLOv5.
YOLOv8 is under active development as of writing this post, as Ultralytics work on new features and respond to feedback from the community. Indeed, when Ultralytics releases a model, it enjoys long-term support: the organization works with the community to make the model the best it can be.

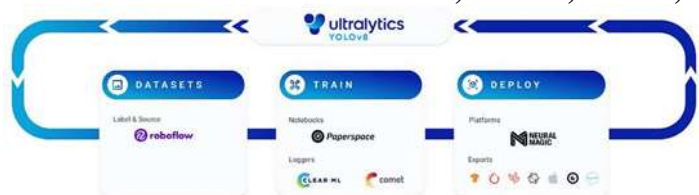

Figure 8: YOLO V8 Process

YOLOv8 Architecture: A Deep Dive

YOLOv8 does not yet have a published paper, so we lack direct insight into the direct research methodology and ablation studies done during its creation. With that said, we analysed the repository and information available about the model to start documenting what's new in YOLOv8.

If you want to peer into the code yourself, check out the YOLOv8 repository and you view this code differential to see how some of the research was done.

Here we provide a quick summary of impactful modeling updates and then we will look at the model's evaluation, which speaks for itself.

The following image made by GitHub user RangeKing shows a detailed visualisation of the network's architecture.

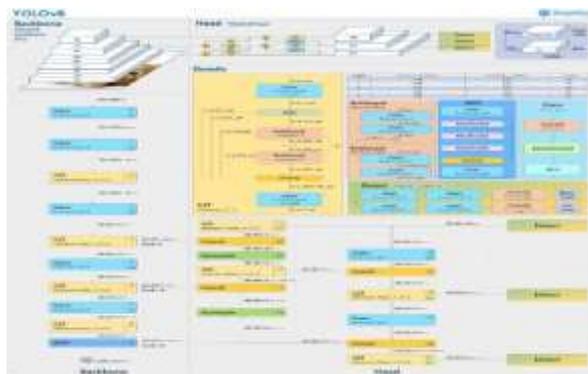

Figure 9 YOLO Architecture

Dataset Description

Kaggle: Here provided a large set of high-resolution retina images taken under a variety of imaging conditions. A left and right field is provided for every subject. Images are labelled with a subject id as well as either left or right.

0    -    No DR
1    -    Mild DR
2    -    Moderate DR
3    -    Severe DR
4    -    Proliferative DR

Kaggle contains many images with poor quality and incorrect labelling.

Training Work -Flow

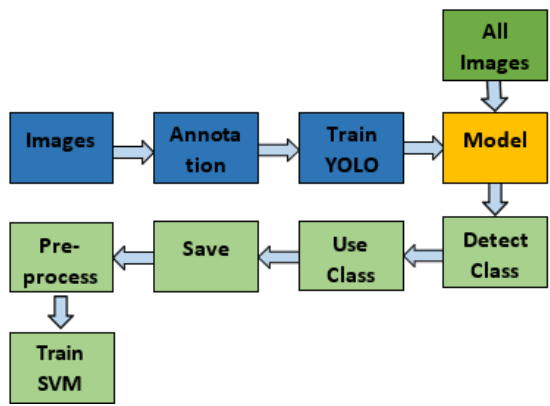

In this approach, we are using some fundus images from all five classes of diabetic retinopathy: normal, mild, moderate, severe, and proliferative. Fundus images are images of the interior surface of the eye that show the retina and its features. We use makesense.ai, which is an online tool that allows us to annotate images by drawing bounding boxes around the objects of interest. We annotate our region of interest (ROI), which is the part of the image that contains the lesions that indicate diabetic retinopathy. Lesions are abnormal changes in the tissue or structure of the eye that can affect the vision. Some common types of lesions are microaneurysms, hemorrhages, exudates, and cotton wool spots.

After completing this task, we pass the image and its annotated labels to the YOLO v8 small model for training. The YOLO v8 small model is a fast and accurate object detection algorithm that can handle multiple classes and scales. It works by dividing the image into a grid of cells and predicting bounding boxes and class probabilities for each cell. By training the model on our annotated images, we can obtain a custom detector that can recognize the objects of interest in new images.

After YOLO training, we apply object detection on the whole diabetic retinopathy image dataset to detect the lesions in the image. We use our custom detector to identify and locate the lesions in each image and mark them by drawing rectangles on them. Along with the detection, we save the type and count of lesions with the image name. By doing this process, we generate a new dataset that is in numerical form. This dataset can be used for further analysis and classification of diabetic retinopathy based on the type and count of the lesions. The type and count of lesions can indicate the severity and progression of diabetic retinopathy and help us to diagnose and treat it accordingly.

After data generation, we normalize the data by applying some pre-processing techniques for machine learning classification models. Machine learning classification models are algorithms that can learn from data and assign labels to new instances based on their features. The pre-processing techniques include scaling, feature selection, dimensionality reduction, and data augmentation. Scaling is a technique that transforms the data to have a standard range or distribution.

It helps to avoid bias and improve convergence in machine learning models. Feature selection is a technique that selects the most relevant and informative features from the data. It helps to reduce noise, redundancy, and complexity in the data and improve performance and accuracy in machine learning models.

Dimensionality reduction is a technique that reduces the number of features in the data by projecting them onto a lower-dimensional space. It helps to avoid overfitting and improve computational efficiency in machine learning models. Data augmentation is a technique that creates new data instances by applying transformations such as rotation, flipping, cropping, or adding noise to the existing data. It helps to increase the diversity and size of the data and improve generalization in machine learning models.

Now our new data is ready to train our SVM model. SVM stands for support vector machine, which is a supervised learning algorithm that can perform classification and regression tasks. We split the data into two subsets: one for training the model and one for testing its performance. We use 80% of the data for training and 20% for testing. We train our SVM model on the training data by finding the optimal hyperparameters that can separate the classes with the maximum margin.

GUI Work-Flow

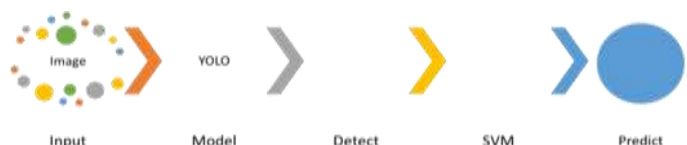
Figure 10 Work Flow

For predicting the new image, we create a user interface by using Tkinter. Tkinter is a Python library that allows us to build graphical user interfaces (GUIs) easily. Our GUI can take an image as input and predict the grade of diabetic retinopathy and show the marked lesions in the image. The grade of diabetic retinopathy indicates the severity of the disease and its impact on the vision.

The GUI works as follows: First, it takes a retinal image as input from the user. Then, it passes the image to the YOLO model that we trained earlier. The YOLO model detects the lesions in the image and marks them by drawing rectangles on them. Next, it counts the number and type of lesions in the image and passes them to the SVM model that we also trained earlier. The SVM model takes the lesion count and type as input and predicts the grade of diabetic retinopathy for the image. Finally, it displays the predicted grade and the marked image to the user.

Result

• Our SVM model Training Score 91% and Testing Score is 84% and our model achieves an Accuracy of 84% % , F1 Score is 81% and Precision is 82% which means that it can correctly predict the grade of diabetic retinopathy upto 84% of the testing images.

• Our YOLO model Accuracy of 78% , F1 Score is 74% and Precision is 72% is our model achieves an, which means that it can correctly Detect the Lesion's of diabetic retinopathy for 78% of the testing images.

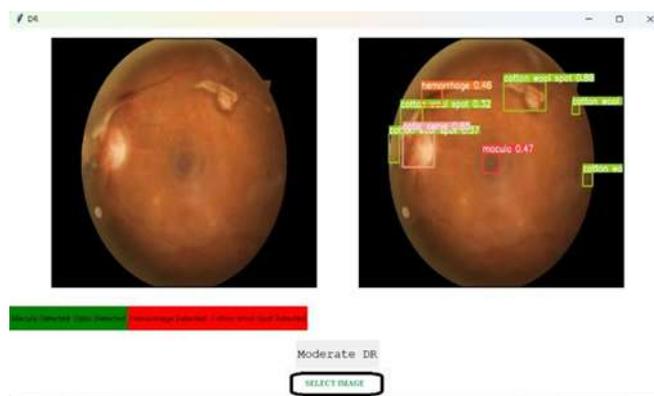

Figure 11:GUI

Conclusion

Using the object detection model YOLO and machine learning classification algorithms, we have achieved promising results in diagnosing diabetic retinopathy with 84% accuracy. This is vital for preventing vision loss in diabetic patients, as timely diagnosis can enable early intervention. Our method can assist ophthalmologists in detecting and classifying diabetic retinopathy more efficiently and effectively, leading to improved patient outcomes. We have test it on our dataset taken from online website yet the clinical test is to be performed. We have test it on our dataset taken from online website yet the clinical test is to be performed.